\documentclass[10pt,twocolumn,letterpaper]{article}
\usepackage[pagenumbers]{cvpr}
\usepackage{microtype}
\definecolor{cvprblue}{rgb}{0.21,0.49,0.74}
\usepackage[pagebackref,breaklinks,colorlinks,allcolors=cvprblue]{hyperref}
\hypersetup{
  pdftitle={Efficient 3D Gaussian Head Avatars for Edge Devices},
  pdfauthor={Umar Farooq, Jean-Yves Guillemaut, Adrian Hilton, Marco Volino}
}

\title{Efficient 3D Gaussian Head Avatars for Edge Devices}

\author{Umar Farooq \quad Jean-Yves Guillemaut \quad Adrian Hilton \quad
Marco Volino\\
CVSSP, University of Surrey}

\begin{document}
\maketitle

\newcommand{\insertFigurePipeline}{%
\begin{figure*}[t]
\centering
\includegraphics[width=\linewidth]{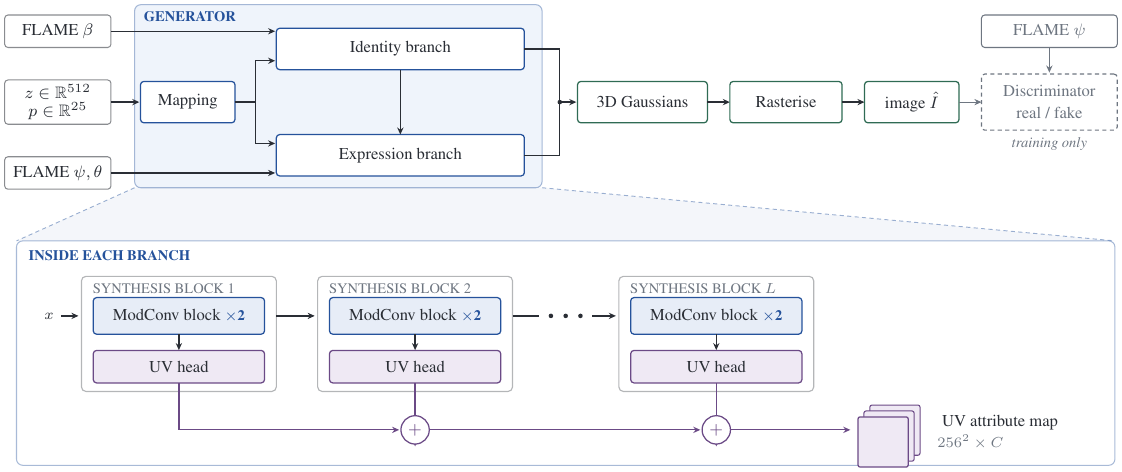}
\caption{Overview of our 3D Gaussian head-avatar generation pipeline.\vspace{-4mm}}
\label{fig:pipeline}
\end{figure*}
}

\newcommand{\insertFigureModConv}{%
\begin{figure}[t]
\centering
\includegraphics[width=\linewidth]{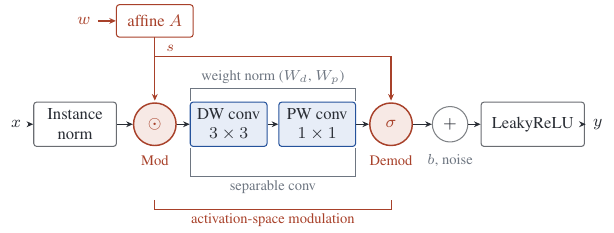}
\caption{Proposed architecture of the modulated convolution block with
  depth-wise separable convolutions. The modulation is applied to the instance
  normalised input feature map and the demodulation to the output of the
  point-wise convolution.\vspace{-4mm}}
\label{fig:mod_conv_block}
\end{figure}
}

\newcommand{\insertFigureBandwidth}{%
\begin{figure}[t]
\centering
\includegraphics[width=\linewidth]{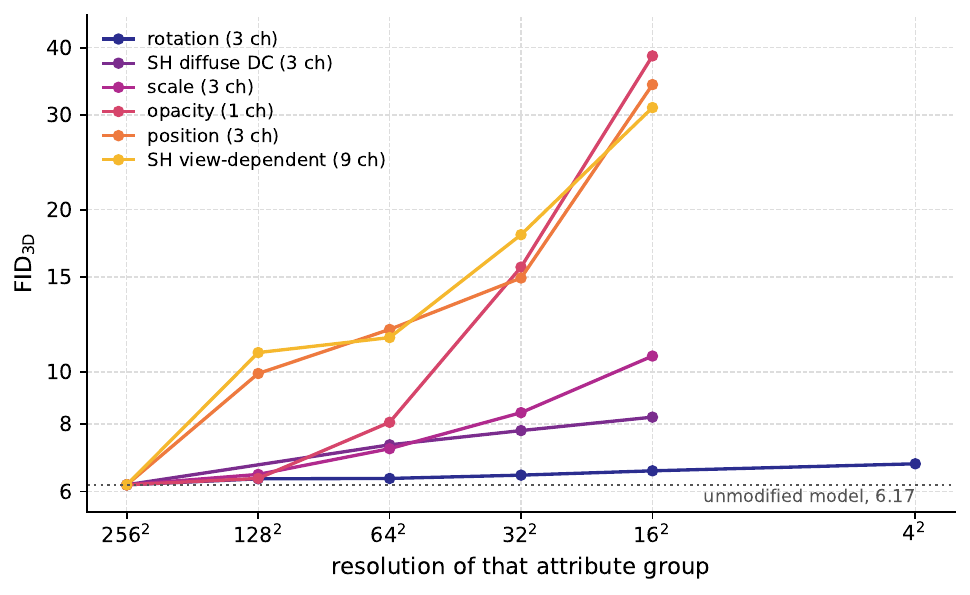}
\caption{Resolution sensitivity of each attribute group. Each curve low-passes
  one group to the resolution on the horizontal axis and bilinearly restores it,
  on frozen weights, leaving the other five groups untouched; the dotted line is
  the unmodified model.}
\label{fig:bandwidth}
\end{figure}
}

\newcommand{\insertFigureQual}{%
\begin{figure}[t]
\centering
\includegraphics[width=\linewidth]{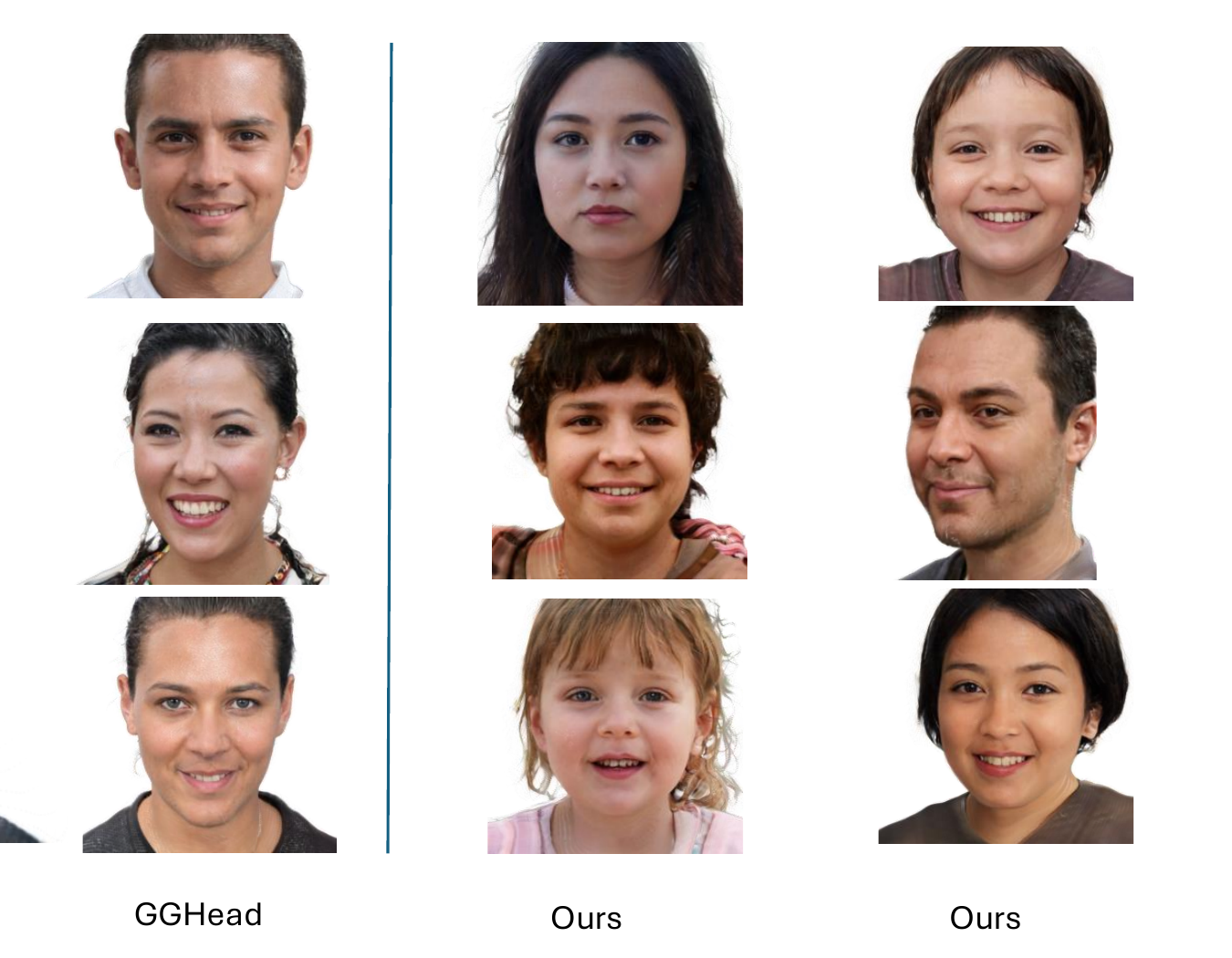}
\caption{Qualitative results from our proposed model and the baseline GGHead
  model at $256^2$ resolution. Please note these are trained for 5M images as
  outlined in Section~\ref{sec:implementation}.}
\label{fig:qual}
\end{figure}
}

\newcommand{\insertFigureQualI}{%
\begin{figure}[t]
\centering
\includegraphics[width=\linewidth]{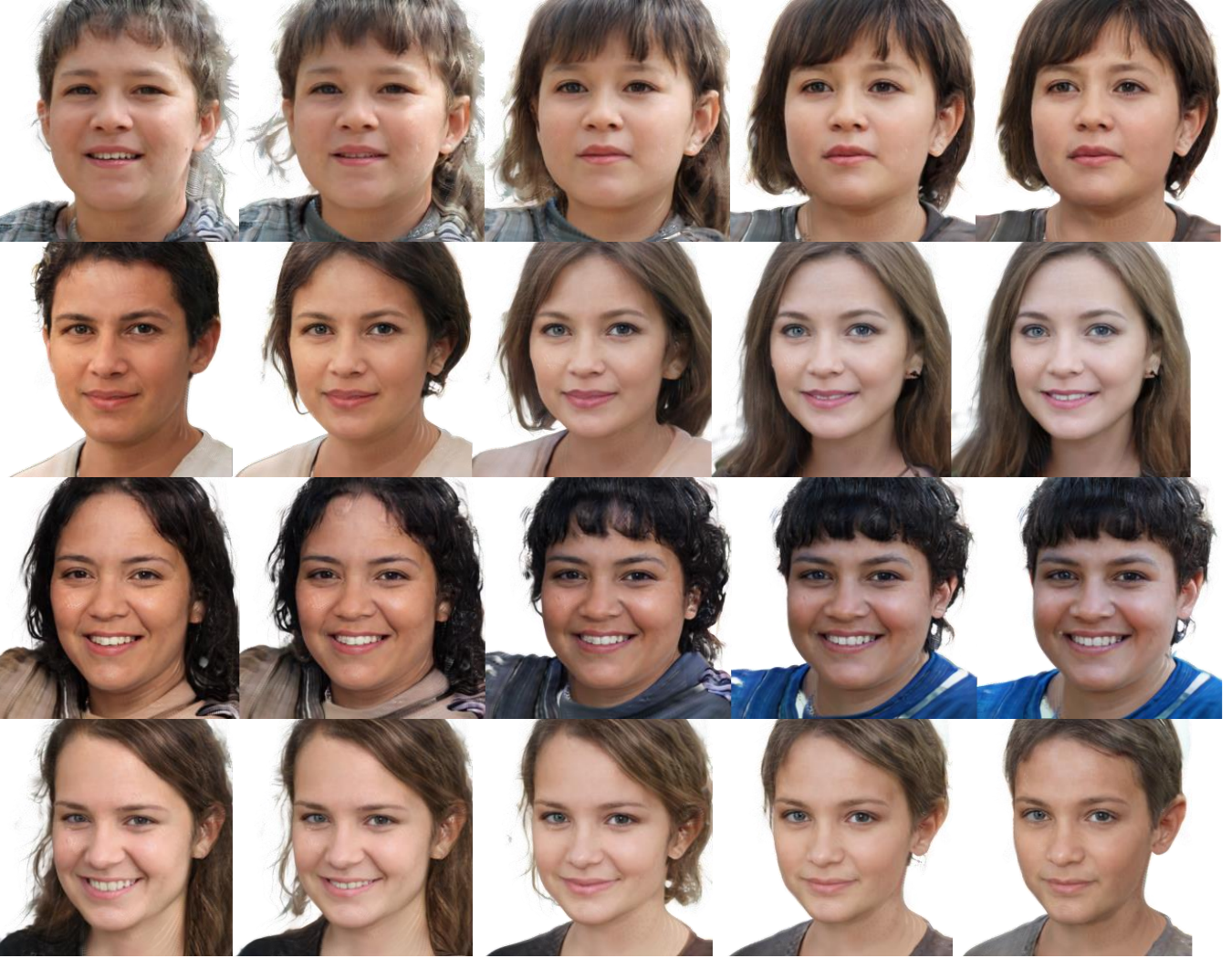}
\caption{Interpolation between random latent codes generated by our static model at $256^2$ resolution. Smooth transitions between identities indicate that meaningful latent interpolation is preserved by the efficient architecture.\vspace{-4mm}}
\label{fig:qualI}
\end{figure}
}

\newcommand{\insertFigurePixelDiff}{%
\begin{figure}[t]
\centering
\includegraphics[width=0.95\linewidth]{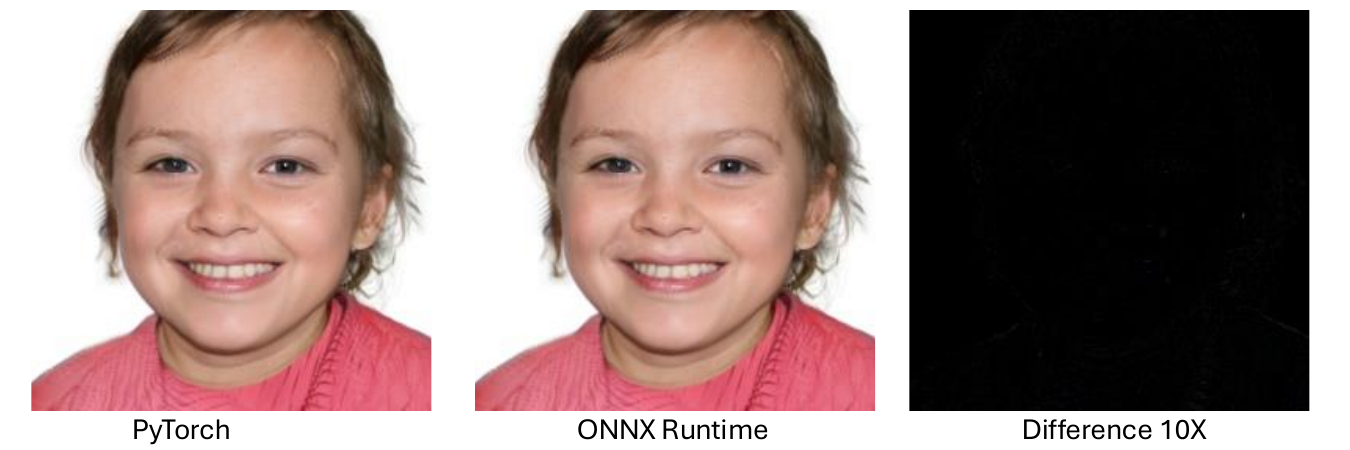}
\caption{A pixel level comparison of our proposed model's inference in PyTorch
  against an onnx-runtime implementation. Please note the pixel differences are
  enhanced by $10\times$ for better visibility.\vspace{-4mm}}
\label{fig:pixel_diff}
\end{figure}
}

\newcommand{\insertFigureAnimation}{%
	\begin{figure*}[t]
		\centering
		\includegraphics[trim=0 10cm 0 0, clip, width=0.9\linewidth]{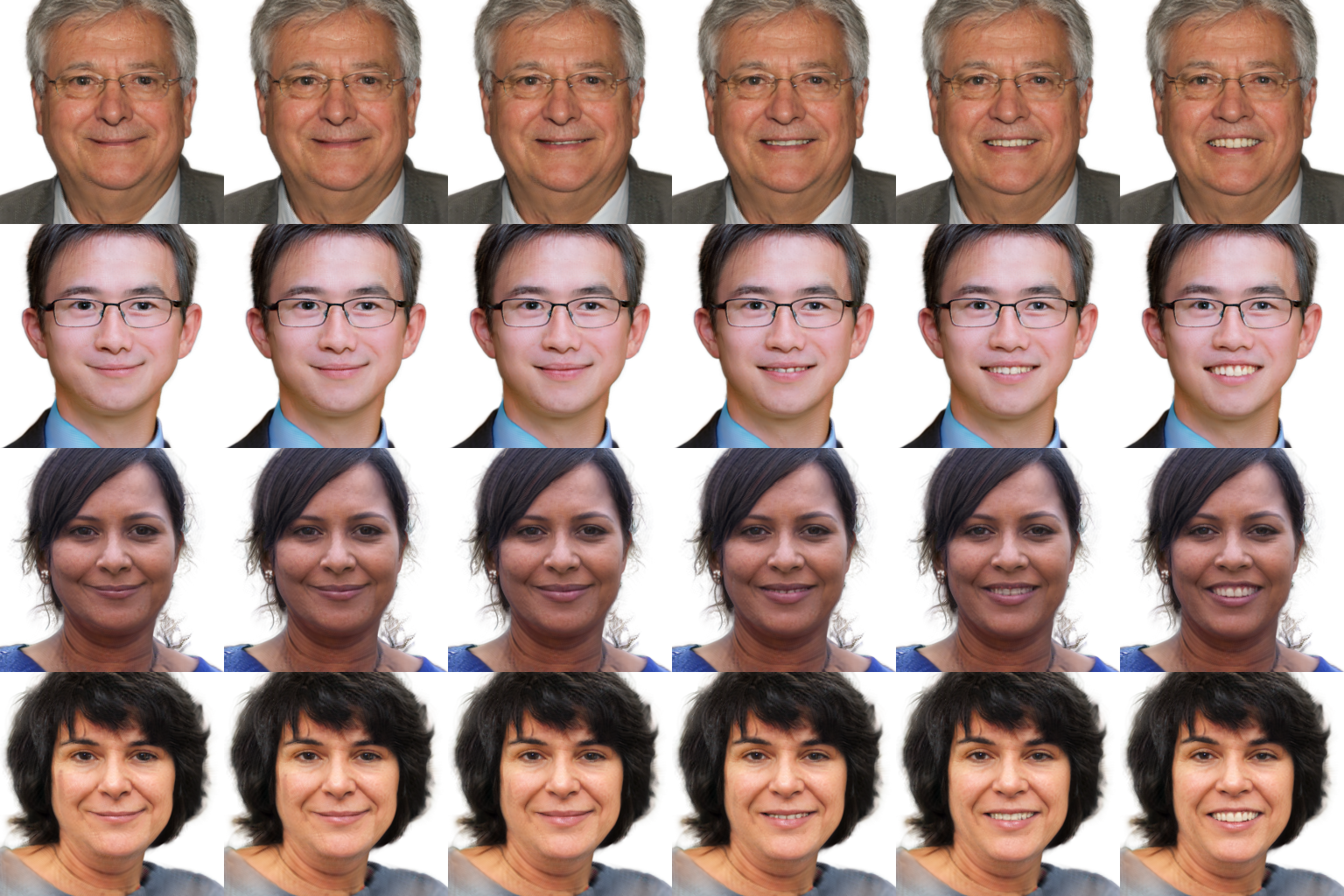}
		\caption{Our animation model can drive avatar expression based on FLAME parameters. In each row the expression is interpolated from the outermost images.\vspace{-4mm}}
		\label{fig:pixel_anim}
	\end{figure*}
}

\newcommand{\insertFigurePaperTeaser}{%
\begin{figure}[tbh]
    \centering
            \centering
            \includegraphics[width=1.0\linewidth]{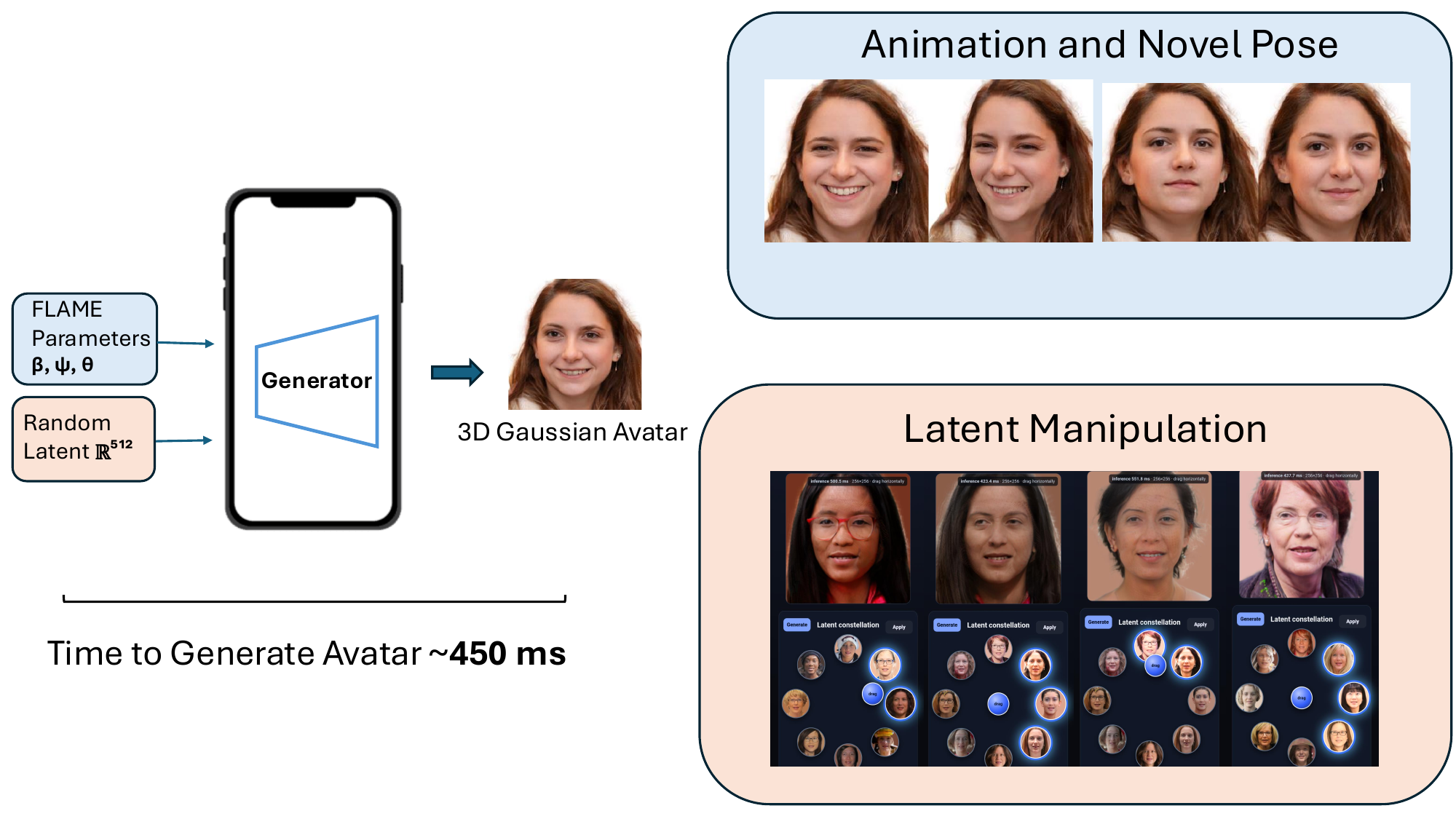}
    \caption{Overview of our efficient 3D Gaussian head-avatar synthesis pipeline and downstream applications. Our model can generate a novel 3D Gaussian avatar from a random latent code and it also supports subsequent animation and latent manipulation.\vspace{-4mm}}
    \label{fig:paper_teaset}
\end{figure}
}
\newcommand{\insertFigureLatentManip}{%
\begin{figure*}[t]
\centering
\includegraphics[width=\linewidth]{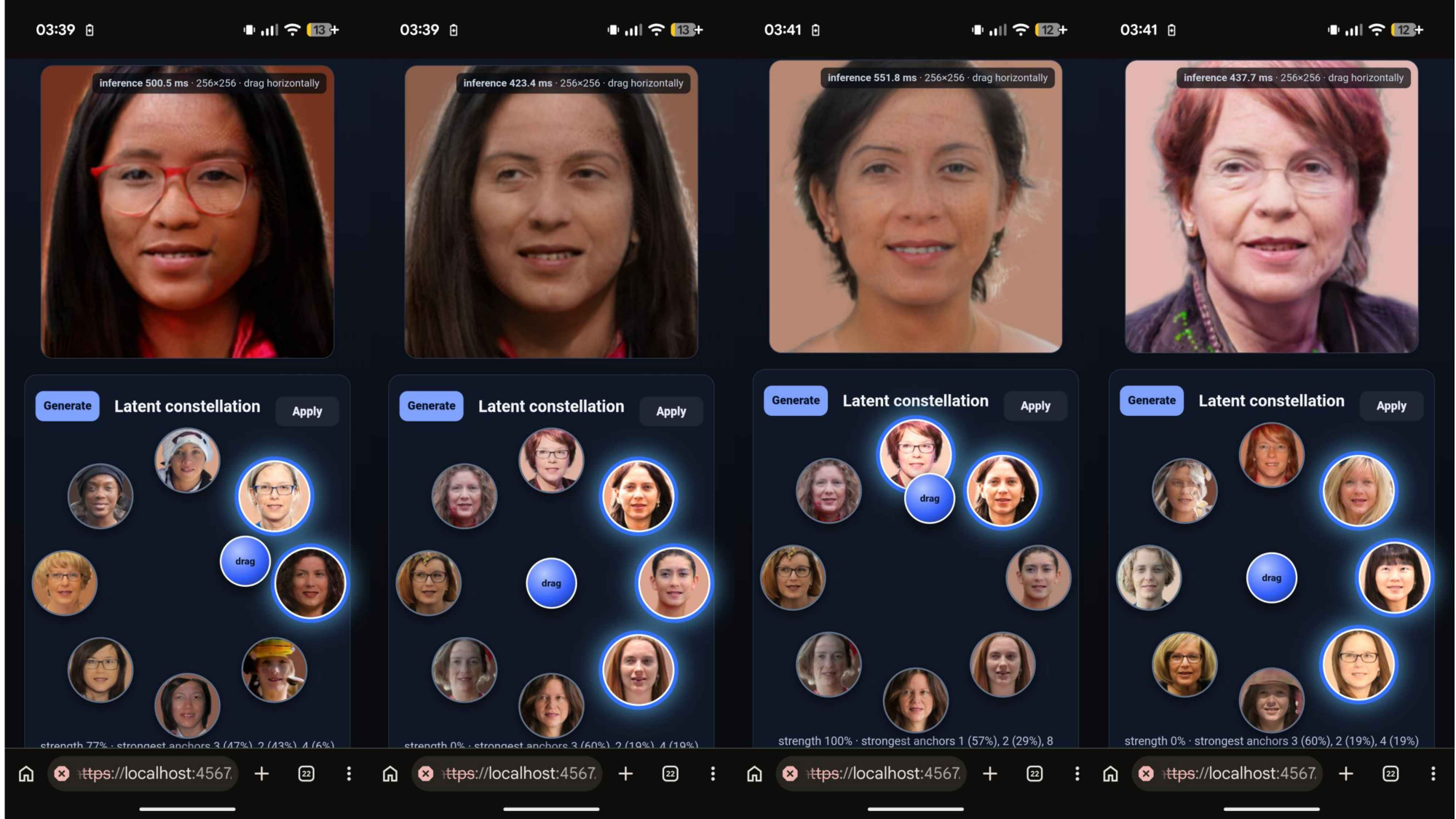}
\caption{Latent space manipulation. Traversing individual directions in the
  intermediate latent space of our generator produces smooth, largely
  disentangled changes to a single factor of variation while the remaining
  identity is preserved, showing that the depth-wise separable synthesis blocks
  retain the editable style space of the baseline architecture.}
\label{fig:latent_manip}
\end{figure*}
}

\newcommand{\insertTableComparisons}{%
\begin{table*}[t]
\centering
\small
\setlength{\tabcolsep}{5pt}
\caption{Comparison of generation quality at $512^2$ rasterisation resolution, together with computational cost, generation latency, and animation latency at 25M training images. FID is computed against $512^2$ reference statistics. The generator backbone, $256^2$ UV attribute map, deformation plane, and Gaussian count are unchanged; consequently, the remaining generator and deployment measurements are the same as for $256^2$ rasterisation.}
\label{tab:comparison}

\begin{tabular}{lccccc@{\hskip 12pt}cc@{\hskip 12pt}c}

\toprule
& \multicolumn{5}{c}{Generation Quality and Compute Cost}
& \multicolumn{2}{c}{Generation}
& \multicolumn{1}{c}{Animation} \\
\cmidrule(lr){2-6}
\cmidrule(lr){7-8}
\cmidrule(lr){9-9}

{Method}
& {FID}$\downarrow$
& {FLOP (G)}$\downarrow$
& {Params (M)}$\downarrow$
& {Train}$\downarrow$
& {Size}$\downarrow$
& {CPU}$\downarrow$
& {Pixel 10}$\downarrow$
& {Pixel 10}$\downarrow$ \\

& & & & {(hrs)} & {(MB)} & {(ms)} & {(ms)} & {(ms)} \\

\midrule

AGORA ($512^2$)~\cite{fazylov2025agora}
& \textbf{3.17}
& 212
& 33
& 50
& 241
& 323
& 2381
& 32 \\

Ours ($512^2$)
& 4.15
& \textbf{13}
& \textbf{10}
& \textbf{29}
& \textbf{46}
& \textbf{148}
& \textbf{453}
& {32} \\

\bottomrule

\end{tabular}
\end{table*}
}

\newcommand{\insertTableMVMetrics}{%
\begin{table}[t] 
	\centering
	\setlength{\tabcolsep}{2pt}
	\scriptsize
    	\caption{We show 3D consistency, multi-view quality and efficiency metrics for our method on a smaller static model at 5M images and $256^2$ resolution.\\[-0.8cm]} 
	\label{tab:combinedmetrics} 
	\begin{tabular}{lccccccc}
		\toprule
		Method & FLOPs & FID & $\text{FID}_{\text{3D}}$$\downarrow$ & $\text{PSNR}_{\text{MV}}$$\uparrow$ & $\text{SSIM}_{\text{MV}}$$\uparrow$  & Train Time$\downarrow$ & VRAM$\downarrow$ \\
		\midrule
		GGHEAD & 66 G & \textbf{9.6} & \textbf{11.57} & 32.60 & 0.868 & \textbf{10.51} hrs. & 21.7 GB \\ 
		Ours-Static & \textbf{2.7} G & 12.0 & 16.40 & \textbf{34.01} & \textbf{0.880} & 11.59 hrs. & \textbf{16.0} GB \\ 
		\bottomrule
	\end{tabular} 

\end{table}
}

\newcommand{\insertTableAblations}{%
\begin{table}[t]
\centering
\caption{Ablations shown at 5M images and $256^2$ resolution. (a) shows that depth-wise
  convolutions are capable of modelling the data distribution and do not lose
  significant quality. Dashes mark configurations we did not profile.}
\label{tab:ablations}
\resizebox{\columnwidth}{!}{%
\begin{tabular}{lcccc}
\toprule
{Configuration} & {FID}$\downarrow$ & {Params}$\downarrow$ &
{FLOP}$\downarrow$ & {CPU}$\downarrow$ \\
 & & (M) & (G) & (ms) \\
\midrule
\multicolumn{5}{l}{\textit{(a) Depth-wise separable convolutions}} \\
GGHead~\cite{kirschstein2024gghead}      & {9.6} & 28.0 & -- & -- \\
\;+ separable synthesis convs            & {9.6} & 8.0  & --   & -- \\
\;+ separable output convs \textbf{(C1)} & 9.8 & 8.0 & -- & 354 \\
\midrule
\multicolumn{5}{l}{\textit{(b) Channel reduction applied to C1}} \\
\;+ feature resolution/ 4                   & 17.2 & 8.0 & -- & 153 \\
\;+ channel width \textbf{(C2)}          & 17.6 & 4.0 & -- & {137} \\
\midrule
\multicolumn{5}{l}{\textit{(c) Spherical harmonics and mixed precision, from C1}} \\
\;+ channel width + SH$_0$               & 11.1 & 4.0 & 7.9 & 197 \\
\;+ feature resolution/ 2                   & 15.8 & 4.0 & {2.7} & 148 \\
\;+ FP16 (last 4 blocks)                 & 12.8 & 4.0 & {2.7} & 158 \\
\bottomrule
\end{tabular}
}
\vspace{-8mm}
\end{table}
}

\begin{abstract}

Generative 3D Gaussian head avatars provide high-quality, efficient rendering, but synthesising the Gaussian representation remains computationally expensive, limiting deployment on resource-constrained and edge devices.
We introduce an efficient generator architecture for unconditional 3D Gaussian head synthesis, based on a parameter-efficient synthesis block and depth-wise separable convolutions while retaining style-based conditioning.
Our architecture reduces generator complexity without requiring model compression or quantisation.
Compared with the baseline model, our approach reduces FLOPs by 94\%, parameter count by 70\%, and model size by 81\%, while maintaining competitive generation quality. 
We further demonstrate practical CPU inference and browser-based execution on mobile devices using ONNX Runtime, enabling 3D Gaussian avatar synthesis without dedicated GPU hardware or application-specific software. 
In addition to conventional image-quality metrics, we evaluate multi-view consistency, training cost, and deployment performance. 
Code, trained models, and evaluation tools will be released publicly.

\end{abstract}

\section{Introduction}
\label{sec:intro}
\color{black}
Generating realistic animatable human avatars is a core research area within computer vision and graphics with extensive downstream applications in the creative industries, gaming, telecommunication, AR/VR and telepresence.
Capturing high-quality human avatars in a studio setting using multiple cameras and depth sensors is expensive and time-consuming.
This has motivated significant research towards automated generation pipelines which can produce high-quality avatars from a single image, multiple images, videos or synthesise novel head identities.

These models rely on large datasets~\cite{Karras_2019_CVPR} to learn useful priors about human geometry, movement and appearance.
However, this is a challenging task and the generated avatars can often lack fine geometric details and multi-view consistency.
\insertFigurePaperTeaser
For producing head avatars with random identities, Generative Adversarial Networks (GANs)~\cite{iangan} have been explored extensively.
~\cite{Chan2021, deng2022gram, voxgraf} use GANs with great success and recently NeRF-based \cite{mildenhall2020nerf} head avatar models have shown good results.
However, NeRF-based representations suffer from slow rendering, which impacts both training and inference.
Recent progress in 3D Gaussian Splatting~\cite{kerbl3Dgaussians} has made it possible to use a differentiable, explicit and fast 3D representation directly inside the training pipeline of a generative model.
~\cite{kirschstein2024gghead} has incorporated 3DGS to produce Gaussian head avatars which are fast to render at both training and inference time, and follow-up work has improved multi-view consistency and background separation~\cite{barthel2026cgsgan} as well as introduced animation capabilities~\cite{yu25gaia, fazylov2025agora}.

The adoption of 3DGS as a differentiable representation has substantially reduced the cost of rasterising avatars; however, generating the Gaussian representation itself remains computationally expensive.
For unconditional head avatar synthesis, the generation pipeline dominates the compute cost and has to be run for every new identity.
Recently, ~\cite{fazylov2025agora} has proposed efficient architectures for the animation pipeline for edge devices, however the initial synthesis step remains costly.
This prevents the use of the full pipeline on resource-constrained and edge devices presenting a bottleneck for various applications.
A lightweight generator capable of producing novel identities in a 3D-consistent manner would enable the complete synthesis pipeline to run on such devices. 
Furthermore, retaining style-based conditioning enables characteristics of the generated head to be modified at runtime, supporting downstream creative and interactive applications.

In this work, we propose a generator architecture that is efficient in terms of both parameter count and computational cost for inference on resource-constrained and edge devices.
We reformulate modulation and demodulation for depth-wise separable convolutions and introduce a parameter-efficient Gaussian attribute output structure while retaining style-based conditioning~\cite{Karras2019stylegan2}.
We demonstrate the performance of our model on commodity hardware and mobile devices inside a web browser with no application-specific software setup required.

\noindent In summary, we make the following contributions:
\begin{itemize}
    \setlength{\itemsep}{0pt}
    \setlength{\parskip}{0pt}
    
    \item We introduce a style-conditioned depth-wise separable synthesis architecture, including a reformulated modulation and demodulation mechanism, together with a parameter-efficient Gaussian attribute output block.
    
    \item We produce a generator which is up to 70\% smaller in parameter count and requires up to 94\% fewer FLOPs, while retaining the original Gaussian output resolution and primitive count.
    
    \item We show that our method can be deployed on commodity hardware and resource-constrained devices, including browser-based inference on mobile phones.
\end{itemize}

\section{Related Work}
\label{sec:related}
\color{black}
\noindent{\bf 3D Aware Generative Head Models:}
3D head avatar synthesis has been an active area of research, with NeRF~\cite{mildenhall2020nerf} and GAN~\cite{iangan} based methods among the most popular approaches.
Approaches using 2D diffusion models with score distillation sampling (SDS) have also been proposed~\cite{HeadSculpt, poole2022dreamfusion}.
\cite{Chan2021} introduced an early approach for direct synthesis of 3D head geometry, adopting the architecture from~\cite{Karras2019stylegan2} and extending it to produce triplanes rendered using NeRF~\cite{mildenhall2020nerf}.
However, this increased the computational cost of both training and inference.
Subsequent work reduced this cost through manifold~\cite{deng2022gram, xiang2023gramhd} or sparse voxel~\cite{voxgraf} representations, by imitating a 2D generator~\cite{mimic3d}, and by extending coverage from the frontal hemisphere to the full head through the untied triplanes of PanoHead~\cite{Pano_2023_CVPR} and spherical parameterisation of SphereHead~\cite{li2024spherehead}.
Next3D~\cite{sun2023next3d} demonstrated that expression, blink and gaze control can be learned from unstructured 2D images without video supervision.
Another line of work uses multi-view diffusion~\cite{tang2024gaf, han2025avatar++, taubner2025cap4d}, but suffers from high computational cost and limitations in multi-view consistency.

Recent work has incorporated 3DGS~\cite{kerbl3Dgaussians} as a replacement for NeRF-based rendering, with significant gains in training and inference speed.
Gaussian Shell Maps~\cite{abdal2024gsm} uses a 2D convolutional generator to predict Gaussian attributes in the texture space of a deformable template, allowing the generator to remain a standard image network while the template provides the geometric prior.
GSGAN~\cite{shen2024gsgan} instead organises the primitives hierarchically and supervises them adversarially.
GGHead~\cite{kirschstein2024gghead} applies the template-UV formulation to head avatars under a FLAME~\cite{FLAME} parameterisation, regularising the geometry while retaining a standard 2D convolutional generator.
CGS-GAN~\cite{barthel2026cgsgan} removes view-conditioned generation that causes multi-view inconsistency in this class of model, reports strong generation quality, and introduces the $\text{FID}_{\text{3D}}$ metric.
3DGH~\cite{kirschstein20253dgh} extends GGHead with separately composable hair and face generators, EGG3D~\cite{li2025egg3d} targets editability, and MVCHead~\cite{chharia2026multi} enforces multi-view consistency without explicit multi-view generation.
Across these approaches, reported efficiency gains primarily concern rendering or training, while the cost of a generator forward pass remains relatively underexplored.
\insertFigurePipeline

\noindent{\bf Animatable Gaussian Head Avatars: }
Animation of Gaussian head avatars is generally achieved by binding primitives to a deformable template.
GaussianAvatars~\cite{qian2024gaussianavatars} follows this mechanism for per-subject reconstruction, while 3D Gaussian Blendshapes~\cite{ma2024gaussianblendshapes} expresses residual deformation in a linear basis, reducing inference-time animation to linear algebra.
GAIA~\cite{yu25gaia} uses a dual-generator architecture, with one generator producing shape and another expression-conditioned deformations, together with a separate expression-conditioned discriminator.
AGORA~\cite{fazylov2025agora} also uses a dual-generator approach but with a single discriminator taking the rendered image and an expression-coded vertex rendering.
It introduces a smaller MLP-based vertex deformation model that approximates the larger expression generator, reducing runtime deformation cost.
Our work instead addresses the synthesis of previously unseen identities and is therefore complementary to these deformation and animation approaches.

\noindent{\bf Efficient Generators and GAN Architectures:}
GAN Compression~\cite{li2020gancompression} introduced channel pruning under distillation for conditional generators, while Content-Aware GAN Compression~\cite{liu2021contentaware} reported an elevenfold reduction in FLOPs for an unconditional StyleGAN2 alongside improved latent disentanglement.
Anycost GANs~\cite{lin2021anycost} vary resolution and channel width at inference to provide multiple operating points from a single model.
Other work directly improves the efficiency of the underlying 2D StyleGAN architecture~\cite{belousov2021mobilestylegan,jia2023blazestylegan}.
MobileStyleGAN~\cite{belousov2021mobilestylegan} uses a wavelet-based generator and parameter-efficient convolutions, removing RGB heads from all but the final synthesis block. It also modulates and demodulates activations rather than weights and pairs depth-wise and point-wise convolutions, but it si based on 2D/wavelet operations.
BlazeStyleGAN~\cite{jia2023blazestylegan} instead retains RGB heads but uses variable-resolution feature maps to reduce computational cost.
Crucially, both MobileStyleGAN~\cite{belousov2021mobilestylegan} and BlazeStyleGAN~\cite{jia2023blazestylegan} have RGB as the final output whereas our method is focused on producing Gaussian output with multiple channels and more structure.

\noindent{\bf Efficient Gaussian Representations:}
A parallel body of work compresses the Gaussians themselves rather than the network that generates them~\cite{bagdasarian20253dgszip}.
Compressed 3DGS~\cite{niedermayr2024compressed} applies sensitivity-aware vector quantisation and entropy coding, LightGaussian~\cite{fan2023lightgaussian} reduces spherical harmonic degree through distillation, and Self-Organizing Gaussian Grids~\cite{morgenstern2024sogs} organise primitives into a two-dimensional grid for standard image codecs.
These methods operate on primitives after optimisation and are complementary to reducing their generation cost.
They could therefore be combined with our approach to further reduce storage requirements.

\noindent{\bf On-Device Inference:}
MobileNeRF~\cite{chen2023mobilenerf} restructures a neural field for execution within the standard rasterisation pipeline of an unmodified browser, while SnapFusion~\cite{li2023snapfusion} and SnapGen~\cite{hu2025snapgen} reduce text-to-image diffusion models to mobile latency budgets.
Measurement conditions are important when interpreting such results: \cite{jia2024webbrowsers} reports that browser-based deep learning inference can be substantially slower than native execution on the same mobile hardware.
Our work builds on GGHead~\cite{kirschstein2024gghead} due to its fast inference~\cite{barthel2026cgsgan} and adoption as a baseline for animation pipelines~\cite{fazylov2025agora, yu25gaia}.
However,~\cite{kirschstein2024gghead, barthel2026cgsgan} focus on performance within conventional GPU-based workflows rather than resource-constrained devices.
AGORA~\cite{fazylov2025agora} improves the animation pipeline for edge devices and demonstrates speed-ups within a web browser, but does not address the cost of initial identity synthesis.
This remains a bottleneck for generating Gaussian head avatars from scratch and prevents on-the-fly modification and style mixing enabled by the StyleGAN2~\cite{Karras2019stylegan2} backbone.
Our work addresses this complementary problem by reducing  memory, compute and storage requirements of initial head avatar synthesis.

\color{black}
\section{Method}
\label{sec:method}

\subsection{Preliminaries}

\noindent{\bf 3D Gaussian Splatting:}
3D Gaussian Splatting~\cite{kerbl3Dgaussians} is an explicit representation for 3D scenes that uses a collection of 3D Gaussians to represent geometry and appearance.
Each Gaussian is defined by its position $\mu$, scale $s$, rotation $r$, opacity $\alpha$, and colour attributes $c$.
The colour $c$ of a pixel is determined by alpha-blending the contributions of $N$ ordered Gaussians overlapping the pixel:
\begin{equation}
c = \sum_{i=1}^{N} c_i \alpha_i \prod_{j=1}^{i-1} (1 - \alpha_j)
\end{equation}
where $c_i$ is the colour of the $i$-th Gaussian and $\alpha_i$ is its opacity at the pixel location.
The term $\prod_{j=1}^{i-1} (1 - \alpha_j)$ represents the transmittance, accounting for the visibility of the $i$-th Gaussian.
View-dependent appearance is represented using spherical harmonics, whose coefficients are evaluated according to the viewing direction.

\noindent{\bf Gaussian Attribute Maps:}
The attributes for each Gaussian are represented as a $C$-channel vector $a = [\mu, r, s, \alpha, c]^\top \in \mathbb{R}^{C}$ in a UV map.
Here $\mu, r, s \in \mathbb{R}^3$ represent position, rotation and scale respectively, $\alpha \in \mathbb{R}$ is the opacity, and $c \in \mathbb{R}^{d_c}$ denotes colour attributes.
We use $C = 13$ ($d_c = 3$) for our final model and $C = 22$ ($d_c = 12$) for the variant incorporating view-dependent effects via first-degree spherical harmonics.
The UV maps are generated at $UV \in \mathbb{R}^{256 \times 256}$ resolution by a StyleGAN2 backbone as a function of a latent code $z$ and a camera pose $p$, where $p \in \mathbb{R}^{25}$ and $z \in \mathbb{R}^{512}$.

\noindent{\bf FLAME Model:}
The FLAME (Faces Learned with an Articulated Model and Expressions) model~\cite{FLAME} is a parametric model of facial shape, expression and pose.
We use FLAME as a geometric prior for the generated Gaussian representation.
The Gaussians are samlped over the mesh surface and a positional zero convolution is used to predict offsets.
The zero convolution allows the Gaussians to be initialised near the mesh surface and to gradually learn their displacement.
This provides a soft geometric regularisation that encourages consistent head geometry across views and poses.
The complete pipeline is shown in Figure~\ref{fig:pipeline}.

\subsection{Architecture Overview}
\noindent{\bf Generator Architecture:}
We start with a StyleGAN2-based generator representing the UV map as a function of camera pose $p$, latent code $z$ and FLAME parameters.
The FLAME parameters are also used by the expression branch and the discriminator, which is shown synthetic geometric cues to improve performance on expressions.
The camera pose and latent code are fed to a mapping network $f(p, z) = w$ where $w \in \mathbb{R}^{512}$.
Here $f$ is a 2-layer MLP with 512 hidden units.
The generator consists of $L$ synthesis blocks, where each block maps
\begin{equation}
F_\ell(\mathrm{Feat}_{\text{in}}, w_\ell, UV_{\ell-1}) = \mathrm{Feat}_{\text{out}}, \mathrm{UV}_{\text{out}}
\end{equation}
Each synthesis block contains a feature layer and a Gaussian output layer whose outputs are accumulated to produce the final UV attribute map.
The Gaussian output layer directly produces the Gaussian attribute channels, with a fixed output resolution for each block.
To reduce the computational and parameter cost of the generator, we replace all normal convolution layers with depth-wise separable convolutions.
A standard convolution $C(W, x)$ can be implemented as a depth-wise convolution followed by a point-wise convolution:
\begin{equation}
\mathcal{C}(W, x) = P(W_p, D(W_d, x))
\end{equation}
where $W_p$ and $W_d$ are the weights for the point-wise and depth-wise convolutions, respectively.
This reduces the convolutional parameter count from:
\begin{equation*}
\begin{aligned}
W = C_{\text{out}} \times C_{\text{in}} \times K \times K
\end{aligned}
\end{equation*}
to:
\begin{equation*}
\begin{aligned}
W_d &= C_{\text{in.d}} \times K \times K \\
W_p &= C_{\text{out.p}} \times C_{\text{in.p}} \times 1 \times 1
\end{aligned}
\end{equation*}
where $K$ is the kernel size and $C_{\text{in.d}}$ and $C_{\text{in.p}}$ are their input channel counts, respectively.

\insertFigureModConv

\noindent{\bf Modulation with Depth-wise Convolutions:}
\label{sec:modulation}
Replacing standard convolutions with depth-wise separable convolutions requires modifying the modulation and demodulation process of the StyleGAN2 generator.
In StyleGAN2, style modulation is applied to the convolution weights.
For a convolution with weights $w_{ijk}$, where $i$ indexes the input channels, $j$ the output channels and $k$ the $K^2$ spatial taps, the style vector $s$ scales the input channels and the weights are then renormalised,
\begin{equation}
w'_{ijk} = s_i \, w_{ijk},
\qquad
w''_{ijk} = \frac{w'_{ijk}}{\sqrt{\sum_{i,k} (w'_{ijk})^2 + \epsilon}} .
\label{eq:demod}
\end{equation}
The normalising sum runs over the input channel axis as well as the spatial axis.
Since different input channels are scaled by different values of $s_i$, no common factor can be taken out of the sum, and the modulation therefore remains dependent on the style vector.

This property does not hold for a depth-wise convolution. 
Each output channel of a depth-wise convolution is produced from a single input channel, so the input and output channel indices coincide, the weights reduce to $w_{jk}$, the style entry acting on channel $j$ is $s_j$, and the sum in Equation~\ref{eq:demod} is taken over $k$ alone,
\begin{equation}
\sigma^{\text{w}}_j
= \sqrt{\textstyle\sum_{k} (s_j w_{jk})^2 + \epsilon}
= \sqrt{s_j^2 \textstyle\sum_{k} w_{jk}^2 + \epsilon} .
\label{eq:sigmaw}
\end{equation}
For small $\epsilon$ the scale can be taken out of the square root and the modulated and demodulated weights become
\begin{equation}
w''_{jk}
= \frac{s_j \, w_{jk}}{|s_j| \sqrt{\sum_{k} w_{jk}^2}}
= \operatorname{sgn}(s_j) \,
  \frac{w_{jk}}{\sqrt{\sum_{k} w_{jk}^2}} .
\label{eq:cancel}
\end{equation}
Thus, except for its sign, the style coefficient $s_j$ cancels during demodulation, removing the magnitude of the style modulation from the depth-wise convolution.
Applying the modulation to the point-wise convolution instead would preserve the style dependence because the $1 \times 1$ convolution restores the sum over the input channels.
However, the depth-wise convolution runs first, so its spatial filtering would remain style-agnostic.
Retaining a dense convolution solely for modulation would also undermine the computational benefit of the separable factorisation.
We therefore apply the modulation to the input activations before the depth-wise convolution, and the demodulation to the output of the point-wise convolution,
\begin{equation}
y = P\big(W_p, \, D(W_d, \, s \odot \tilde{x})\big),
\qquad
\hat{y}_j = \frac{y_j}{\sigma_j},
\label{eq:block}
\end{equation}
where $\tilde{x}$ is the instance normalised input and $\odot$ is a per-channel scaling.
The demodulator $\sigma_j$ is the norm of the composed separable kernel that produces output channel $j$, taken over the input channels $i$ and the spatial taps $k$ of the point-wise and depth-wise weights and scaled by the style,
\begin{equation}
\sigma_j = \sqrt{\textstyle\sum_{i,k}\big(s_i\, W_{p,ji}\, W_{d,ik}\big)^2 + \epsilon} .
\label{eq:sigma}
\end{equation}
Because the point-wise stage restores the sum over $i$, no common factor can be taken out of Equation~\ref{eq:sigma}, and the cancellation of
Equation~\ref{eq:cancel} does not recur.
The style therefore remains present in the composed depth-wise/point-wise operation while retaining the efficiency of separable convolution.

Weight demodulation was originally introduced to avoid normalizing the activations, which produced droplet artifacts in StyleGAN~\cite{Karras2019stylegan2}, so moving back to the data domain loses that property.
To stabilise the adversarial training, we apply weight normalisation~\cite{salimans2016weightnorm} to the convolution weights and instance normalisation~\cite{ulyanov2016instancenorm} to the input feature maps.
The bias and noise are added after demodulation and before the LeakyReLU activation function.
Removing either normalization destabilised training in our experiments.
The architecture of our modulated convolution block is shown in Figure~\ref{fig:mod_conv_block}.

\noindent{\bf Template Mesh Mapping:}
From the Gaussian attributes in the UV map, a set of 3D Gaussians is sampled for the rendering process.
We use a layer on the produced $\mu$ with zero weight initialisation to initially place Gaussians exactly on the surface similar to~\cite{kirschstein2024gghead}.
This provides geometric regularisation and encourages consistent geometry across views and poses.

\subsection{Generator Design}
\label{sec:allocation}
To further reduce FLOPs, we reduce the width of the intermediate feature channels to one-quarter and one-half of their original size for our static model and our main animatable model.
We also reduce the maximum number of channels in each layer to 256 for our static model, further reducing the computational cost. 
For the static model, we further reduce the number of channels produced by the Gaussian output heads by decreasing the number of spherical harmonic coefficients.
The output Gaussian attributes are still generated at the original spatial resolution to ensure the image resolution and Gaussian count remain unchanged.
Maintaining the UV map dimensions allows these reductions to be evaluated independently of the number of generated Gaussian primitives.

Finally, in our static model, we train the last four synthesis blocks in
FP16~\cite{micikevicius2018mixed} precision instead of FP32.
Besides the reduction in memory requirements, these design changes also improve generation quality at a fixed parameter budget, as demonstrated in the ablation study in Section \ref{sec:ablations}.

\subsection{Loss Function}
We use the non-saturating GAN loss~\cite{iangan}:
\begin{equation}
L_{\text{adv}} = \mathrm{softplus}(-D(R(G),p,\psi))
\end{equation}
where $D$ and $G$ are the discriminator and generator respectively, $R$ is the renderer, $p$ is the camera pose and $\psi$ is the FLAME expression code.
The generator loss is:
\begin{equation}
L_G = L_{\text{adv}} + \lambda_p L_{\text{pos}} + \lambda_s L_{\text{scale}}
\end{equation}
where the position and scale regularisation terms are defined as:
\begin{equation}
L_{\text{pos}} = \left\lVert M_{\text{pos}} \right\rVert_{2}, \quad
L_{\text{scale}} = \left\lVert M_{\text{scale}} \right\rVert_{2}
\end{equation}
and $M_{\text{pos}}$ and $M_{\text{scale}}$ are the predicted position offset and predicted scale, respectively.

\color{black}
\section{Results and Evaluation}
\label{sec:results}
\label{sec:exp}
\noindent{\bf Dataset:}
We use the FFHQ dataset~\cite{Karras_2019_CVPR} for training and evaluation.
The dataset consists of 70k high-quality images of human faces with a wide variety of pose, expression, and lighting conditions.
MODNet~\cite{MODNet} is used to remove the background and isolate the head region for training our model.

\noindent{\bf Evaluation Metrics and Baselines:}
We use FID, FLOPs, inference latency and file size as our primary evaluation metrics.
FID~\cite{heusel2017gans} is used to measure the visual quality of the generated head avatars, while FLOPs and inference latency measure the computational cost of our model at runtime.
File size measures the storage and transfer requirements of the model, which are important for deployment on resource-constrained devices.

For our main results in Table~\ref{tab:comparison}, we use 50K real and generated samples for FID calculation.
For the ablation studies in Table~\ref{tab:ablations} and Table~\ref{tab:combinedmetrics}, we calculate FID using 10K generated samples and 10K real samples from FFHQ.
We also report $\text{FID}_{\text{3D}}$~\cite{barthel2026cgsgan} with 10K images, which measures generation quality under large deviations from the conditioning view.
The heads are conditioned on a frontal view while the rendering viewpoints are randomly sampled from the dataset, exposing view-conditioning bias and geometric inconsistency.

We measure $\text{PSNR}_{\text{MV}}$ and $\text{SSIM}_{\text{MV}}$ following the protocol of~\cite{xiang2023gramhd} and GGHead~\cite{kirschstein2024gghead}, generating 30 multi-view images of 100 random heads, and reconstructing the surface using~\cite{neus2}.
These multi-view metrics measure agreement between renderings of the same identity across viewpoints.
We compare our proposed approach against AGORA~\cite{fazylov2025agora}, which adds an expression deformation branch to the same synthesis backbone.
Our static generator is compared against~\cite{kirschstein2024gghead} which is the baseline adopted by various methods in this subdomain like~\cite{yu25gaia,fazylov2025agora}.

\noindent{\bf Implementation Details:}
\label{sec:implementation}
The proposed generator block is used throughout the model. 
The block used to predict the Gaussian attributes is a modified version of our proposed generator block with noise disabled. 
We train our model on 8 NVIDIA H200 GPUs.
We use the Adam optimizer with learning rates of 0.0025 and 0.002 for the generator and discriminator, respectively, and a batch size of 32.
We set $\lambda_p=0.1$ and $\lambda_s=0.05$, apply R1 gradient regularisation to the discriminator with a weight of 1, and use minibatch standard deviation regularisation in the discriminator.
We train our model to produce UV maps at $256^2$ resolution and cap the number of Gaussians at 65K. 
Not all Gaussians are active for each avatar as their contribution is controlled by the predicted opacity.

We train the first stage of our model for 6.5M images and then on an additional 15M images overall in the second stage. %
For the ablation runs reported in Table~\ref{tab:ablations}, we train for 5M images. 
The model is trained in PyTorch and the final checkpoint is exported to ONNX for inference.
We calculate FID on the PyTorch checkpoints for all experiments.
Figure~\ref{fig:pixel_diff} shows that both ONNX and PyTorch outputs are similar when using the same renderer.

For computational-cost evaluation, we use onnx-tool~\cite{ThanatosShinji} to parse and profile the exported ONNX models. 
CPU inference latency is measured using ONNX Runtime capped to 4 cores on an AMD EPYC 9535 CPU. 
Mobile results are obtained using ONNX Runtime Web out of the box with a Pixel 10 smartphone. 
Further details on the cost evaluation and the full measurement protocol are provided in the supplementary material.

For our main model, we train an additional deformation branch similar to ~\cite{fazylov2025agora} and use our proposed synthesis blocks in both the synthesis and deformation branches. 
Unless otherwise stated, images are rasterised at $256^2$. For the full-resolution
comparison in Table~\ref{tab:comparison}, we rasterise the same predicted
Gaussians at $512^2$; the UV attribute map, deformation plane, Gaussian count,
and generator architecture remain unchanged at $256^2$.
We report runtime separately for initial Gaussian generation and subsequent expression-conditioned animation in Table~\ref{tab:comparison} to distinguish the contribution of the synthesis and deformation stages.
We use the MLP proposed by~\cite{fazylov2025agora} to speed up animation with new expressions.
More details are provided in the supplementary material.

\subsection{Quantitative Comparison}
\insertTableComparisons
\insertFigurePixelDiff
\insertFigureAnimation

Our proposed method achieves a 70\% reduction in parameter count and a 94\% reduction in FLOPs compared to the baseline model, as shown in Table \ref{tab:comparison}. 
The storage size for our model also decreases by 81\%, as shown in Table \ref{tab:comparison}, making it easier to store, transfer, and deploy on edge devices.
The reported file sizes use the uncompressed, unquantised model weights, leaving further potential reductions through compression and quantisation.
In preliminary experiments, we found that some parameters are more sensitive to quantisation than others. 
We therefore leave quality-aware quantisation for future work.
Further details are provided in the supplementary material. 
Our model maintains competitive FID while substantially reducing the computational and storage requirements of the generator.

At $512^2$ rasterisation resolution, our model obtains an FID of 4.15 compared
with 3.17 for AGORA. These values are computed against $512^2$ reference
statistics. Because the generator still predicts a $256^2$ UV attribute map
and the $512^2$ setting changes only rasterisation, the FLOPs, parameter count,
training time, exported model size, and generator and animation latencies in
Table~\ref{tab:comparison} are unchanged. Appendix
Section~\ref{sec:suppl_comparison} provides the full architecture and cost
breakdown.

\noindent{\bf Synthesis Latency:}
Our model reduces CPU inference latency by 54\% compared with the baseline, making it more suitable for deployment on edge devices. 
However, the reduction in latency is smaller than the reduction in FLOPs. 
We attribute this in part to ONNX Runtime execution of modulated convolution and upsampling operations as combinations of standard operators, together with the overhead of the additional upsampling operations used to maintain the UV map resolution. 
Custom ONNX operators and further runtime optimisation could therefore provide additional latency reductions.
Our model runs the entire unconditional face synthesis pipeline, including the MLP mapping network, in 148 ms on a CPU. 
This can be combined with a lightweight 3DGS renderer to support complete avatar generation and rendering on edge devices.
For our phone evaluation we render the Gaussians on device, but this uses a different renderer with a reduced feature set than what the model was trained with.
Table \ref{tab:comparison} also reports browser-based mobile inference on a Google Pixel 10, using ONNX Runtime Web without GPU acceleration or runtime-specific optimisation.
These results demonstrate that the proposed generator can execute on CPU and mobile hardware without requiring dedicated GPU acceleration. 
\insertTableMVMetrics

\noindent{\bf Multi-view Consistency:}
\label{sec:training_cost}
Table \ref{tab:combinedmetrics} reports multi-view consistency and resource requirements for our static generator. 
Our method reduces peak GPU memory by 26\%, from 21.7 GB to 16.0 GB, under identical testing conditions. 
The proposed model also improves $\mathrm{PSNR}_{\mathrm{MV}}$ from 32.60 to 34.01 and $\mathrm{SSIM}_{\mathrm{MV}}$ from 0.868 to 0.880, indicating improved agreement between renderings of the same identity across viewpoints. 
However, $\mathrm{FID}_{\mathrm{3D}}$ increases from 11.57 to 16.40. 
This discrepancy suggests that the metrics capture different aspects of multi-view generation.%
It can possibly point to some over smoothing happening in our method resulting in this dichotomy.
The experiment indicates an increase in training time but it only runs for 5M images and our full run on the larger model shown in Table~\ref{tab:comparison}. reduces overall training time by 42\%.

\subsection{Ablations}
\label{sec:ablations}
Table \ref{tab:ablations} isolates the contribution of the architectural and efficiency modifications introduced in Section \ref{sec:method}.
Replacing the standard synthesis convolutions with our depth-wise separable blocks preserves the baseline FID of 9.6 while reducing the parameter count from 28M to 8M.
Replacing the Gaussian output convolutions produces only a small increase in FID to 9.8, demonstrating that the proposed separable architecture can substantially reduce model capacity without significantly degrading generation quality. 
We next evaluate whether similar efficiency gains can be obtained through direct capacity reduction. 
Reducing the feature resolution decreases CPU latency from 354 ms to 153 ms, while further reducing the channel width reaches 137 ms and 4M parameters.
However, FID increases substantially to 17.2 and 17.6, respectively, showing that naive downscaling alone produces an unfavourable quality-efficiency trade-off.
Finally, we evaluate spherical harmonic reduction, feature resolution and mixed-precision training from the C1 configuration. 
Reducing the channel width and spherical harmonic degree to SH$_0$ achieves an FID of 11.1 with 4M parameters and 197 ms latency. 
Further reducing the feature resolution lowers the computational cost to 2.7G FLOPs and latency to 148 ms, although FID increases to 15.8. 
Training the final four synthesis blocks in FP16 recovers generation quality to an FID of 12.8 at the same 2.7G FLOPs, with a small increase in latency to 158 ms.
Together, these results show that the efficiency gains of the final architecture do not arise from capacity reduction alone.
The proposed separable blocks preserve generation quality under substantial parameter reduction, while the additional architectural and precision choices provide complementary trade-offs between quality, model size and inference latency.

\subsection{Qualitative Results}
\insertFigureQualI
Figure~\ref{fig:qualI} shows interpolations between random latent codes generated by our proposed model at $256^2$ resolution.
The resulting sequences exhibit smooth transitions between identities without abrupt visual discontinuities, indicating that the efficiency modifications preserve meaningful interpolation within the learned latent space.
This is particularly important for our architecture because retaining style-based modulation enables continuous control of the generated Gaussian attributes despite the use of depth-wise separable convolutions.
Additional qualitative comparisons with the baseline and further generated examples are provided in the supplementary material.
Figure~\ref{fig:paper_teaset} also shows our method learns meaningful degrees of variation. We provide further analysis in the supplementary material.

\noindent{\bf Animation Results:}
We additionally apply our proposed synthesis blocks to an animatable generator to demonstrate their applicability beyond static head synthesis.
Figure~\ref{fig:pixel_anim} shows our animation model can produce expression-conditioned deformations from FLAME parameters, including independent eye and jaw motion.
Further results are provided in the supplementary material, including animation driven by FLAME parameters estimated from monocular video.

\insertTableAblations
\section{Conclusion}
\label{sec:conclusion}
\noindent{\bf Summary: }
We introduce an efficient generator architecture that substantially reduces
the computational requirements of 3D Gaussian head synthesis while retaining
competitive generation quality.
Our approach reduces parameter count by 70\% and computational cost by 94\%
compared with the baseline, while reducing model size by up to 81\%.
Crucially, these gains enable the complete synthesis of new 3D Gaussian head
avatars on CPUs and mobile devices through a web browser, without
requiring dedicated GPU acceleration or application-specific software.
Beyond efficient rendering or re-animation of existing avatars, our approach
makes the generation of new identities practical on resource-constrained
hardware, addressing a key bottleneck in existing 3D Gaussian avatar pipelines.
The proposed architecture therefore provides a practical foundation for
deployable generative 3D avatar systems, enabling interactive synthesis and
personalisation across edge, mobile and web-based applications.

\noindent{\bf Limitations: }
The reduction in inference latency is smaller than the corresponding reduction in computational cost, indicating that practical performance remains constrained by the available runtime operations and their implementation.
We also observe a discrepancy between the improved $\text{PSNR}_{\text{MV}}$ and $\text{SSIM}_{\text{MV}}$ metrics and the degradation in $\text{FID}_{\text{3D}}$, which requires further investigation.
Reducing the spherical harmonic representation also limits the ability of the most efficient model to represent view-dependent appearance.
The tested implementation did not make use of WebGPU~\cite{WebGPU}, which could provide hardware-accelerated browser execution.
Future work will investigate sensitivity-aware quantisation and optimised
operators for convolution and modulation to further reduce model size and
inference latency.

\noindent{\bf Ethical Considerations: }
The FFHQ dataset used to train our models contains demographic biases that
may be reflected in the distribution and quality of the generated avatars.
Reducing the computational requirements for generating realistic head
avatars also lowers the barrier to their creation and therefore introduces
dual-use risks, including their potential use for misinformation and
deepfakes.
We therefore encourage the development and adoption of appropriate
safeguards alongside continued research into efficient generative avatar
technology.

{
    \small
    \bibliographystyle{ieeenat_fullname}
    \bibliography{main}
}

\clearpage
\appendix

\section{Implementation and Evaluation Details}
\label{sec:suppl_protocol}

\subsection{Training and Measurement Protocol}
\label{sec:suppl_training}
\label{sec:suppl_measurement}
The results for the CPU inference time are obtained by using the onnx-runtime in
a Python environment with an AMD EPYC 9535 CPU, with the runtime restricted to
four threads and four CPU cores for all reported results, including
Table~\ref{tab:comparison} of the main paper.
We report the best run over repeated measurements after a warm-up pass.
For the edge device evaluation we leave the inter- and intra-process thread
counts up to the runtime environment and manually restrict them on the desktop
CPU.
The device is kept out of the power saving mode and background processes are
minimized to ensure the best possible performance.
Note that on the edge device the model runs without any GPU acceleration.
The reported model sizes are the file sizes on disk, without any additional
compression applied, which leaves open the scope for further space savings and
faster network transfer.
All FID and $\text{FID}_{\text{3D}}$ numbers are computed on the PyTorch
checkpoints rather than on the exported ONNX checkpoints, since the export is
used only for the efficiency measurements; Section~\ref{sec:suppl_onnx} shows
that the two produce near-identical outputs, so the quality numbers transfer to
the exported model.

\subsection{Computational Cost Evaluation}
\label{sec:suppl_cost}
The edge device cost is measured with the onnx-runtime on a Pixel 10 mobile
phone.
Both the full generation pipeline and the animation MLP are run and benchmarked
individually.
The model is served inside a simple web page and the inference and the
evaluation are performed entirely on the mobile device.
All dependencies are also sent over the network, so we additionally report the
overall transfer size.
Transferring the full model, comprising the weight files, the SVD basis of the
animation network, the onnx runtime and the FLAME data, amounts to
$75.59$\,MiB across $24$ files in total, which is quick to transfer to a mobile
device over a stable internet connection, and restarting the application is
considerably faster once these files are cached.
For FLOP calculation we report the Forward MACs reported by
onnx-tool~\cite{ThanatosShinji}, where one MAC is defined as
1 MAC = float(a) * float(b) + float(c).
The reported parameter and FLOP counts cover the exported generator only, and
therefore exclude the discriminator, which is not used at inference time, and
the Gaussian rasteriser, which is not part of the exported graph.

\section{Architecture for Full Resolution}
\label{sec:suppl_comparison}
Table~\ref{tab:comparison} reports FID at $512^2$ rasterisation resolution while
retaining the generator and deployment measurements of the corresponding
$256^2$ setting. Here we provide the complete architecture and cost breakdown
that explains why those measurements are unchanged.
It is important to note that this is the same network as the $256^2$ model: the
UV attribute backbone still ends at a $256^2$ UV map and the deformation plane
resolution is likewise still $256^2$.
Moving to a $512^2$ output changes the resolution at which the Gaussians are
rasterised, not the network that produces them.
Consequently the structural cost of the generator is unchanged, and the
parameter count, the FLOP count and the exported file size at $512^2$ are
identical to the values measured for the $256^2$ model. The exported generator
graph is also identical; differences between latency figures reported from
separate profiling runs reflect measurement conditions rather than a change in
the network.
The additional cost of rendering at $512^2$ therefore lies entirely in the
Gaussian rasteriser, which remains a native CUDA implementation outside the
exported graph, and the numbers in Table~\ref{tab:suppl512} should not be read
as the cost of producing a $512^2$ output.

Table~\ref{tab:suppl512} reports the breakdown for this model.
The full generator contains $10{,}673{,}119$ parameters, requires
$12.735$\,GFLOP, equivalently $6.367$\,GMAC, per forward pass and exports to an
onnx graph of $46.116$\,MB.
On the desktop CPU described in Section~\ref{sec:suppl_measurement} the median
latency of the full pipeline is $246.8$\,ms on a single core, and $171.3$\,ms
when four threads are used.
The synthesis and deformation branches are also profiled individually, and the
deformation branch accounts for roughly a fifth of the parameters while
contributing a substantially larger share of the FLOPs, since it operates on
the full deformation plane.
We verify the export at this configuration as in Section~\ref{sec:suppl_onnx},
and the maximum relative deviation between the PyTorch and onnx outputs is
$2.99\times 10^{-6}$, well inside our acceptance threshold of $10^{-4}$.

Table~\ref{tab:suppl256fid} reports the corresponding generation quality at
$256^2$ resolution, complementing the $512^2$ results in
Table~\ref{tab:comparison} of the main paper.
Our model reaches an FID of $4.3$ against $3.8$ for
AGORA~\cite{fazylov2025agora}.
The gap in quality is modest given that our generator is the same network used
for $512^2$ rasterisation, and it is obtained at the substantially reduced
parameter, FLOP and storage budget reported in Table~\ref{tab:suppl512}.

\begin{table}[t]
\centering
\footnotesize
\setlength{\tabcolsep}{3pt}
\caption{Cost breakdown of the generator used for $512^2$ rendering. The
  branches are profiled individually while the full model is timed as a single
  end-to-end run of the complete pipeline, so the branch latencies do not add up
  directly to the full-model latency. All reported times are medians on a single
  CPU core; the full model takes $171.3$\,ms when four threads are used.
  Parameter, FLOP and size figures are identical to those of the $256^2$ model,
  since the UV attribute backbone and the deformation plane both remain at
  $256^2$ and only the rasterisation resolution changes.}
\label{tab:suppl512}
\begin{tabular}{@{}lrrrr@{}}
\toprule
Component & Params (M) & GFLOP & Size (MB) & Time (ms) \\
\midrule
Synthesis   & 7.74  & 7.42  & 31.51 & 134.8 \\
Deformation & 2.13  & 5.32  & 14.63 & 129.4 \\
\midrule
Full        & 10.67 & 12.74 & 46.12 & 246.8 \\
\bottomrule
\end{tabular}
\end{table}

\begin{table}[t]
\centering
\footnotesize
\caption{Generation quality at $256^2$ resolution. FID is computed on the
  PyTorch checkpoints against reference statistics at the matching resolution,
  following the protocol of Section~\ref{sec:suppl_measurement}.}
\label{tab:suppl256fid}
\begin{tabular}{@{}lr@{}}
\toprule
Method & FID $\downarrow$ \\
\midrule
AGORA~\cite{fazylov2025agora} & 3.8 \\
Ours                          & 4.3 \\
\bottomrule
\end{tabular}
\end{table}

\section{Architecture Details}
\label{sec:suppl_ablation}

\subsection{Depth-wise Separable Convolutions}
\label{sec:suppl_depthwise}
We replace all the convolutions in the synthesis layer with depthwise
convolutions for the first experiment in Table~\ref{tab:ablations}(a) of the main
paper but use normal convolutions in the UV channel synthesis layer.
For the second experiment we replace all the convolutions in the synthesis layer
and the UV channel synthesis layer with depthwise convolutions.
There is a significant reduction in the parameter count however there is minimal
change in the FID score of the models.

\subsection{Spherical Harmonics, Channel Width and Precision}
\label{sec:suppl_sh_precision}
For the spherical harmonics ablation we start with the DWBase configuration,
halve the number of channels and reduce the spherical harmonics degree to 0.
In the second experiment we also halve the channel height and width on top of the
previous changes, and in the final experiment we also train the last 4 synthesis
blocks in FP16 precision instead of the normal FP32 precision.
Halving the channel width and height makes the model faster than the baseline by
36\% but it suffers in terms of quality as indicated by the FID score, and we
hypothesize that the reduced capacity in the model forces a loss in terms of the
data distribution modelling capability.
Using FP16 precision for the last 4 synthesis blocks potentially allows the model
to store and analyze more combinations of features in the earlier layers with
smaller spatial dimensions.
For the later layers with higher spatial dimensions the solution space is smaller
and the combinations can be spread more spatially.
The inference time however increases slightly, which could be due to the runtime
cost of converting FP16 tensors to FP32 for the operations which are not
supported in FP16 precision.
We also convert the UV maps back to FP32 before rendering them and this could be
contributing to additional load.

\subsection{Generator and Feature Map Sizes}
\label{sec:suppl_arch}
Each synthesis block includes two of the modulated convolution blocks of
Figure~\ref{fig:mod_conv_block} together with a Gaussian output layer whose
outputs are accumulated into the final UV map.
We use different size feature maps inside individual synthesis blocks to further
reduce parameters and FLOPs.
Additional upsampling operations are used before the Gaussian output layer to
ensure a consistent UV map output resolution and therefore an unchanged number of
Gaussians across all of our variants.
Keeping the Gaussian count fixed ensures that the comparisons in the main paper
reflect the capacity of the generator and not a change in the number of
primitives being rendered.

\subsection{Generator Output Heads}
\label{sec:suppl_negative}
We also experiment with removing the cumulative creation of the Gaussian output
channels following~\cite{belousov2021mobilestylegan} but it has an adverse impact
on the quality and fidelity of the generated avatars similar
to~\cite{jia2023blazestylegan}.
We posit it removes direct gradient propagation to earlier resolution synthesis
blocks which is critical for stable training especially with Gaussian attributes
which are very sensitive to minor changes.
A small error in a predicted position channel displaces a primitive in 3D where
it is amplified by the renderer, whereas the same error in an RGB generator only
perturbs a single pixel value.
We therefore retain the cumulative structure in our generator.

\section{Quantisation Sensitivity}
\label{sec:suppl_quantisation}
The file sizes reported in Table~\ref{tab:comparison} of the main paper are for
uncompressed and unquantised FP32 weights, so they represent an upper bound on
the storage cost of our model.
In preliminary experiments we applied uniform post-training quantisation to the
generator weights and observed that the sensitivity is highly uneven across the
network.
Some layers tolerate reduced precision with little change in FID, whereas the
Gaussian output heads, and in particular the layers producing the position and
scale channels, degrade quickly.
This mirrors the behaviour discussed in Section~\ref{sec:suppl_negative}: an
error in a position channel displaces a primitive in 3D and is amplified by the
renderer, while the same error in a colour channel perturbs only the appearance
of one primitive.
A uniform quantisation of the whole generator therefore trades a substantial
part of the quality for the additional size reduction, while leaving the
Gaussian attribute heads in higher precision retains quality at a smaller
saving.
A quality-aware scheme that assigns precision per attribute group is the natural
next step, and we leave it for future work rather than reporting a partial
result here.

\section{ONNX Export Ablation}
\label{sec:suppl_onnx}
Figure~\ref{fig:pixel_diff} of the main paper shows the pixel level difference
between the output of our model in PyTorch and the onnx-runtime
implementation~\cite{microsoft}.
The onnx-runtime implementation is able to closely replicate the output from the
PyTorch implementation with minimal pixel differences, which are enhanced
$10\times$ in that figure for better visibility.
Please note that during the conversion process some highly optimized operations
like modconv are re-implemented using a combination of standard supported
operations, so this check confirms that the efficiency numbers in the main paper
are measured on a model equivalent to the one evaluated for quality.
The residual differences are concentrated at high-frequency edges and are
consistent with the accumulation of floating point differences between the fused
PyTorch kernels and their decomposed ONNX equivalents.

For the onnx-runtime~\cite{microsoft} based evaluation we produce the UV maps
from the onnx model after conversion from PyTorch, and the rendering is then
performed using the original Python and PyTorch codebase with the FLAME template
mesh to ensure consistency with the quality evaluation.
On the mobile phone the Gaussians are instead rasterised by a web based splat
renderer, which differs from the renderer used during training and introduces
minor visual artifacts that are not present in the PyTorch results.

\section{Additional Qualitative Results}
\label{sec:suppl_qualitative}
\insertFigureQual
Figure~\ref{fig:qual} compares samples from our proposed model against the
baseline GGHead model at $256^2$ resolution, with both models trained for 5M
images.
The identities produced by our model are of comparable visual quality despite
the substantially reduced parameter count and FLOP budget, and the geometry
remains stable across the sampled identities.
Further generated examples are included in the supplementary video.

\section{Analysis of Learned Variation}
\label{sec:analysis}

\subsection{Latent Space Variation}
\label{sec:suppl_variation}
\insertFigureLatentManip
Figure~\ref{fig:paper_teaset} of the main paper shows that our model learns
meaningful degrees of variation, and we analyse this further here.
Because we retain style-based modulation in the depth-wise separable synthesis
blocks, the generator keeps an intermediate latent space in which individual
directions correspond to interpretable factors of variation.
Figure~\ref{fig:latent_manip} shows the effect of traversing such directions:
moving along a single direction changes one factor while leaving the remaining
identity largely intact.
The transitions are smooth and free of abrupt discontinuities, which indicates
that the reduced-capacity generator has not collapsed onto a small set of modes
and that the mapping from the latent space to the Gaussian attribute maps
remains locally well behaved.
This matters for our setting because the manipulation is applied to the style
codes and therefore costs a single additional forward pass through the
generator, so identity editing remains available on an edge device without any
optimisation loop at inference time.
We observe that directions affecting appearance, such as colouring and
illumination, are more cleanly separated than directions affecting geometry,
which is consistent with the attribute sensitivity analysis in
Section~\ref{sec:bandwidth}, where the geometric attribute groups are the ones
carrying the highest resolution requirement.

\subsection{Gaussian Attribute Sensitivity}
\label{sec:bandwidth}
\insertFigureBandwidth
The generator emits position, rotation, scale, opacity and spherical harmonic
coefficients as a single map at a single resolution, an arrangement inherited
from image synthesis, where the three output channels are samples of one
radiometric field and a shared sampling rate is therefore appropriate.
The channels of a Gaussian attribute map are unrelated physical quantities, and
we do not assume that they share a resolution requirement.
We estimate the requirement of each group independently by low-pass filtering
that group to a reduced resolution, restoring it bilinearly to the full
resolution, and leaving the remaining groups unmodified, which yields the
marginal value of resolution to that group alone.
All measurements are taken at inference time on frozen weights and are reported
in $\text{FID}_{\text{3D}}$, since the intervention acts on geometry.

The result is shown in Figure~\ref{fig:bandwidth}, and the groups behave very
differently.
The position channels are by far the most sensitive: reducing their resolution
degrades $\text{FID}_{\text{3D}}$ well before any other group is affected,
because a low-pass filter applied to positions moves neighbouring primitives
towards a common location and directly removes geometric detail that the
renderer then cannot recover.
Scale and rotation follow, but with a visibly flatter response, since these
attributes vary more slowly across the surface of the head and neighbouring
primitives are already similar.
The opacity and spherical harmonic groups are the most tolerant and can be
reduced substantially before the curve departs from the unmodified model, which
is the behaviour expected of quantities that are smooth over the UV map.
Two observations follow.
First, the common practice of emitting every attribute at a single shared
resolution over-provisions the appearance channels in order to satisfy the
requirement of the position channels, and this over-provisioning is paid for on
every forward pass.
Second, the ordering of the curves matches the quantisation behaviour reported
in Section~\ref{sec:suppl_quantisation}, where the position and scale heads were
also the least tolerant to a reduction in precision.
Both results point at the same conclusion, that the geometric attributes carry
the information budget of the representation, and that a generator which spends
its resolution and its precision per attribute group rather than uniformly is
the more promising direction for further efficiency gains.

\section{Additional Animation Results}
\label{sec:suppl_animation}
Figure~\ref{fig:pixel_anim} of the main paper shows that our animation model
produces expression-conditioned deformations from FLAME parameters, including
independent eye and jaw motion.
Further animation results are included in the supplementary video, including
avatars driven by FLAME parameters estimated from monocular video.

\end{document}